\documentclass[11pt]{article}

\usepackage[preprint]{acl}

\usepackage{times}
\usepackage{latexsym}
\usepackage{amsmath}
\usepackage{amssymb}
\usepackage{dsfont}

\usepackage{booktabs}
\usepackage{multirow}
\usepackage{makecell}
\usepackage{adjustbox}
\usepackage{mathtools}
\usepackage{algorithmic}
\usepackage{algorithm}
\newcommand\blfootnote[1]{%
  \begingroup
  \renewcommand\thefootnote{}\footnote{#1}%
  \addtocounter{footnote}{-1}%
  \endgroup
}
\usepackage[T1]{fontenc}

\usepackage[utf8]{inputenc}

\usepackage{microtype}

\usepackage{inconsolata}

\usepackage{graphicx}
\usepackage{tcolorbox}

\title{\textsc{CuMA}: Aligning LLMs with Sparse Cultural Values via \\ Demographic-Aware Mixture of Adapters}

\author{
\textbf{Ao Sun\textsuperscript{1}},
\textbf{Xiaoyu Wang\textsuperscript{1}},
\textbf{Zhe Tan\textsuperscript{1}},
\textbf{Yu Li\textsuperscript{1}},
\textbf{Jiachen Zhu\textsuperscript{2}},
\textbf{Yuheng Jia\textsuperscript{1,3}$^\ast$},
\textbf{Shu Su\textsuperscript{1}$^\ast$} \\
\textsuperscript{1}Southeast University \\
\textsuperscript{2}ByteDance Inc. \\
\textsuperscript{3}Key Laboratory of New Generation Artificial Intelligence Technology \\
and Its Interdisciplinary Applications (Southeast University), Ministry of Education, China \\
\texttt{\{sunao, yhjia, sushu\}@seu.edu.cn}
}

\begin{document}
\maketitle
\blfootnote{$^\ast$ Corresponding author.}

\begin{abstract}
As Large Language Models (LLMs) serve a global audience, alignment must transition from enforcing universal consensus to respecting cultural pluralism. We demonstrate that dense models, when forced to fit conflicting value distributions, suffer from \textbf{Mean Collapse}, converging to a generic average that fails to represent diverse groups. We attribute this to \textbf{Cultural Sparsity}, where gradient interference prevents dense parameters from spanning distinct cultural modes. To resolve this, we propose \textbf{\textsc{CuMA}} (\textbf{Cu}ltural \textbf{M}ixture of \textbf{A}dapters), a framework that frames alignment as a conditional capacity separation problem. By incorporating demographic-aware routing, \textsc{CuMA} internalizes a \textit{Latent Cultural Topology} to explicitly disentangle conflicting gradients into specialized expert subspaces. Extensive evaluations on WorldValuesBench, Community Alignment, and PRISM demonstrate that \textsc{CuMA} achieves state-of-the-art performance, significantly outperforming both dense baselines and semantic-only MoEs. Crucially, our analysis confirms that \textsc{CuMA} effectively mitigates mean collapse, preserving cultural diversity. Our code is available at \url{https://github.com/Throll/CuMA}.
\end{abstract}

\section{Introduction}

\begin{figure*}[t]
\centering
\includegraphics[width=0.95\textwidth]{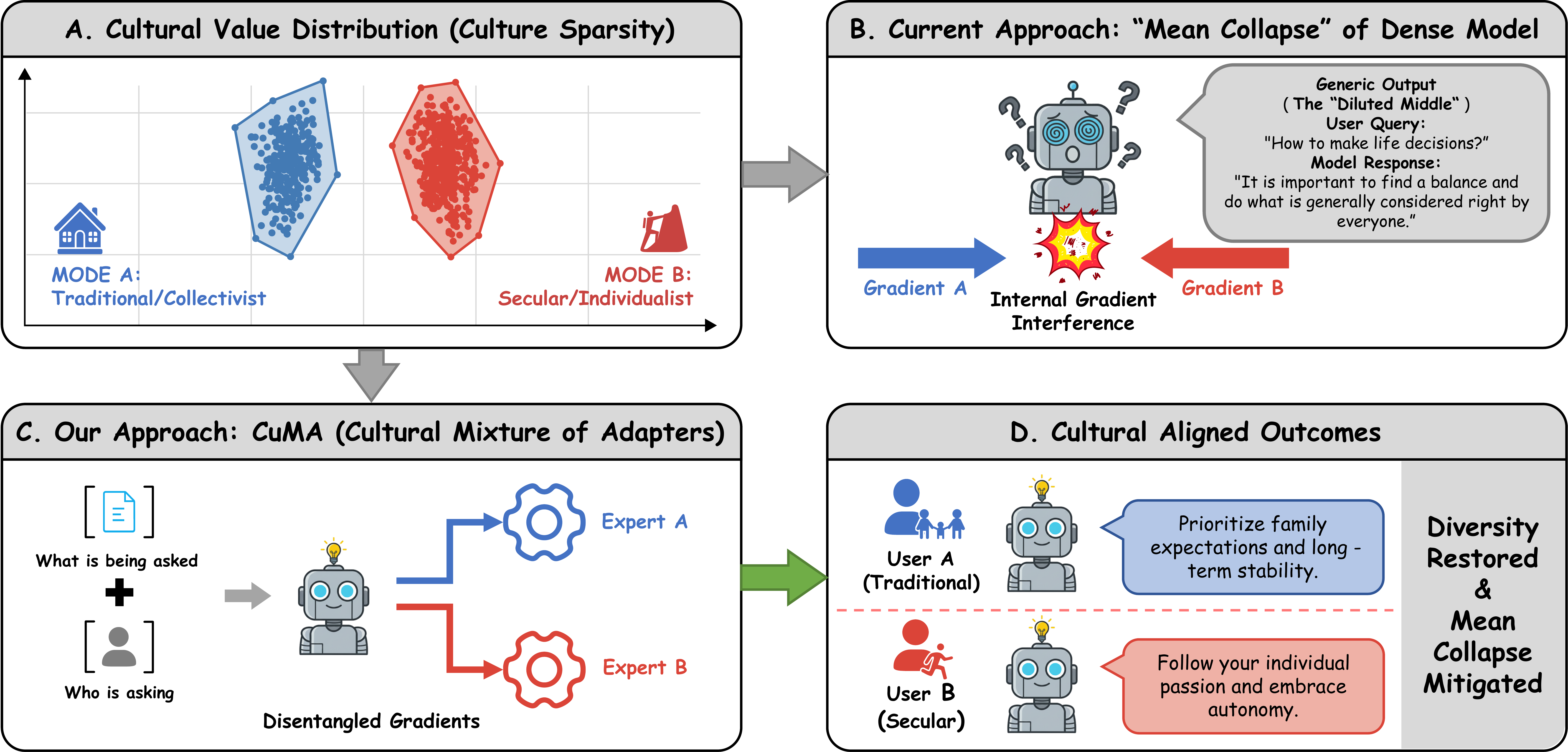} 
\caption{\textbf{Mechanism of Mean Collapse and the \textsc{CuMA} Solution.}
(A) Human values exhibit \textit{Cultural Sparsity}, forming distinct modes (e.g., Traditional vs. Secular). 
(B) Standard dense models suffer from \textbf{Gradient Interference} when optimizing for conflicting modes simultaneously. This forces the model into \textbf{Mean Collapse} (the "Diluted Middle"), producing generic responses that fail to resonate with any group. 
(C) \textbf{\textsc{CuMA}} addresses this via \textit{Demographic-Aware Routing}, explicitly disentangling gradients into specialized experts. 
(D) Consequently, the model generates distinct, culturally resonant outcomes for diverse users, effectively restoring value diversity.}
\label{fig:teaser}
\end{figure*}

Large Language Models (LLMs) have achieved remarkable success in general-purpose reasoning~\cite{gao_large_2024}. To ensure these models remain helpful and harmless, alignment techniques like Reinforcement Learning from Human Feedback (RLHF)~\cite{christiano2017deep, ouyang2022training} are widely adopted. This paradigm typically uses a monolithic reward model to capture human preferences~\cite{Frick:EECS-2025-82}. This approach is effective for consensus-based tasks, such as safety compliance~\cite{ijcai2024p0586}, code generation~\cite{chen2021evaluatinglargelanguagemodels}, and mathematical reasoning~\cite{zhang-etal-2025-lessons}, where a globally optimal response generally exists.

However, as LLMs serve a global user base, alignment must extend to cultural resonance~\cite{adilazuarda-etal-2024-towards, oh-etal-2025-culture}. In subjective domains, response utility is culturally contingent, meaning a response considered insightful in one community may be irrelevant in another~\cite{khamassi_strong_2024}. Consequently, human values are inherently pluralistic and often conflicting~\cite{sorensen2024roadmappluralisticalignment}. Existing methods~\cite{christiano2017deep, ouyang2022training, NEURIPS2023_a85b405e, gu2025infifpoimplicitmodelfusion} optimize a dense set of parameters over such data, implicitly assuming a unified value system. When minimizing error across conflicting modes, dense models gravitate towards a statistical average, leading to Mean Collapse.

This results in the model collapsing divergent values into a single dominant representation, suppressing minority perspectives and imposing a monolithic consensus~\cite{durmus2024measuringrepresentationsubjectiveglobal}. Mean Collapse manifests as "mode-covering" behavior, where models output generic, diluted responses. Crucially, this average is rarely neutral. Driven by imbalances in pre-training corpora~\cite{alkhamissi-etal-2024-investigating,zhu2025toremitopicawaredatareweighting,oncel-etal-2024-adaptation} and the homogeneity of crowd-sourced annotators~\cite{li-etal-2025-assessing,10447803}, the learned "mean" often reflects Western, Educated, Industrialized, Rich, and Democratic (WEIRD) norms~\cite{santurkar2023whose,Henrich_Heine_Norenzayan_2010}. 

We argue that this failure is rooted in gradient interference. Human values are inherently sparse~\cite{KOSTINA20151019}, clustering into distinct, conflicting modes rather than forming a continuous spectrum~\cite{LIU2025104099}. A single dense model cannot simultaneously fit these opposing clusters~\cite{sukiennik2025evaluationculturalvaluealignment,adilazuarda-etal-2024-towards}. Consequently, to minimize global error, it converges to a statistical average, or the "diluted middle", as visualized in Figure~\ref{fig:teaser}.

To address this, we propose \textsc{CuMA}, a framework that reformulates alignment as a conditional capacity separation problem. Standard Mixture-of-Experts (MoE) route tokens based solely on internal hidden states~\cite{10.5555/3600270.3600785,li2024mixtureofexpertsllmsecretlyembedding}, struggling to distinguish culturally conflicting preferences within similar contexts~\cite{wang-etal-2024-scaling}. This design is motivated by the insight that cultural differences are driven by both semantic and demographic proxies~\cite{adilazuarda-etal-2024-towards}. Therefore, \textsc{CuMA} conditions expert selection on the joint representation of semantic content and the user's demographic profile. This allows the router to learn a latent cultural topology, where parameter subspaces are specialized not just by what is being asked, but by who is asking, effectively isolating gradients and preserving cultural diversity~\cite{fu2022latenttopologyinductionunderstanding}.

Our contributions are as follows: (1) We formally identify cultural sparsity as the geometric root of alignment failure in pluralistic settings, demonstrating that dense parameterization inevitably leads to \textit{Mean Collapse}, a structural inability to resolve conflicting modes;

(2) We propose \textsc{CuMA}, a framework that implements conditional capacity separation via demographic-aware routing to explicitly disentangle conflicting gradients into specialized parameter subspaces, allowing the model to learn a latent cultural topology that isolates interference;

(3) Extensive evaluations on WorldValuesBench, Community Alignment, and PRISM show that \textsc{CuMA} achieves state-of-the-art performance, significantly outperforming dense baselines. Analysis confirms that this disentanglement effectively restores generative diversity and mitigates the Mean Collapse found in standard dense models.

\section{Problem Formulation}

In this section, we establish the theoretical foundations of our framework. From a probabilistic perspective, we formulate cultural alignment as a conditional modeling task dependent on demographic context. We then characterize the geometry of pluralistic values through the lens of \textit{Cultural Sparsity}, and analyze why dense parameterization fails to capture this geometry, leading to \textit{Mean Collapse}.

\subsection{Cultural Alignment as Conditional Modeling}

We formalize cultural alignment as a conditional modeling problem, where response validity depends on the user's cultural context. Let $\mathcal{X}$ denote the space of inputs (e.g., prompts), $\mathcal{Y}$ the space of responses, and $\mathcal{D}$ the set of demographic profiles (e.g., region, ideology) serving as observable proxies for latent cultural values. The objective is to learn a conditional model $P_\theta(y \mid x, d)$ that maximizes the likelihood of culturally resonant responses.

A crucial distinction exists between \textit{consensus} and \textit{non-consensus} tasks. For consensus-based tasks (e.g., safety~\cite{lu2025alignmentsafetylargelanguage,10.5555/3780338.3783444} or math reasoning~\cite{ahn-etal-2024-large,azerbayev2024llemmaopenlanguagemodel}), an optimal response $y^*$ is invariant to user attributes (i.e., $P(y|x, d) \approx P(y|x)$), and a single model is sufficient. However, cultural alignment is inherently a \textit{non-consensus} problem~\cite{tao_cultural_2024}: the optimal response distribution varies across $\mathcal{D}$, and conflicting values can be simultaneously valid. To maximize utility, the model should explicitly model the dependency on $d$, rather than marginalizing over it.

\subsection{Cultural Sparsity}

While distinct cultures often share universal commonalities, their preference distributions in the latent representation space typically exhibit multimodal structures, where divergent value systems form separate clusters. We term this geometric property \textit{Cultural Sparsity}.

\textbf{Definition 2.1 (Cultural Sparsity).} Let $P^*(y \mid x, d_i)$ and $P^*(y \mid x, d_j)$ be the conditional value distributions for two distinct demographic profiles. Let $\mu_k \in \mathbb{R}^m$ and $\Sigma_k \in \mathbb{R}^{m \times m}$ denote the mean vector and covariance matrix of group $k$ in the $m$-dimensional representation space (where $m$ is the ambient dimensionality, e.g., the hidden size of the LLM). Defining the pooled covariance as $\bar{\Sigma}_{ij} = \frac{1}{2}(\Sigma_i + \Sigma_j)$, we categorize the distributions as \textit{culturally sparse} if the Mahalanobis distance between their centers significantly exceeds the ambient dimension $m$:
\begin{equation}
\label{eq:sparsity}
(\mu_i - \mu_j)^\top \bar{\Sigma}_{ij}^{-1} (\mu_i - \mu_j) \gg m
\end{equation}

This inequality implies that inter-group divergence dominates intra-group dispersion. Under such sparsity, a single dense representation is geometrically incapable of covering disjoint modes simultaneously. Consequently, the model collapses diverse values into a single expectation, failing to accurately capture distinct cultural preferences (see Appendix~\ref{sec:appendix_density_gap}).

\subsection{The Failure of Dense Models: Mean Collapse}

Standard alignment methods optimize a dense model $P_\theta(y \mid x)$ by minimizing the forward Kullback-Leibler (KL) divergence $D_{\text{KL}}(P_{\text{data}} \parallel P_\theta)$. While the distribution of models is theoretically complex~\cite{machina-mercer-2024-anisotropy}, the shared parameterization across conflicting groups forces the model to capture the central tendency of the aggregate gradient. We analyze this behavior using a unimodal proxy in the representation space.

\textbf{Theorem 2.1 (Mean Collapse).} Under the assumption of cultural sparsity (Eq. \ref{eq:sparsity}), consider a dense estimator $P_\theta$ constrained to a single-component exponential family (e.g., a Gaussian) with mean parameter $\mu_\theta$. The solution minimizing the forward-KL divergence satisfies $\mu_\theta^* = \mathbb{E}_{P_{\text{data}}}[y]$, converging strictly to the global mixture mean. Consequently, the model exhibits \textit{mode-covering} behavior: it centers its probability mass in the "diluted middle", a solution that is statistically optimal for minimizing global error, yet fails to capture the inherent plurality of cultural values. We provide comprehensive derivations in Appendix~\ref{sec:appendix_derivations}: Appendix~\ref{sec:appendix_optimization} proves the mean-matching property; Appendix~\ref{sec:appendix_density_gap} quantifies the exponential density decay at the collapsed mean; Appendix~\ref{sec:appendix_variance} demonstrates the resulting variance inflation; and Appendix~\ref{sec:appendix_resolution} theoretically establishes the resolution via conditional routing.

\section{\textsc{CuMA}: Modeling Latent Cultural Topology via Conditional Routing}

\begin{figure}[t]
\centering
\includegraphics[width=0.95\columnwidth]{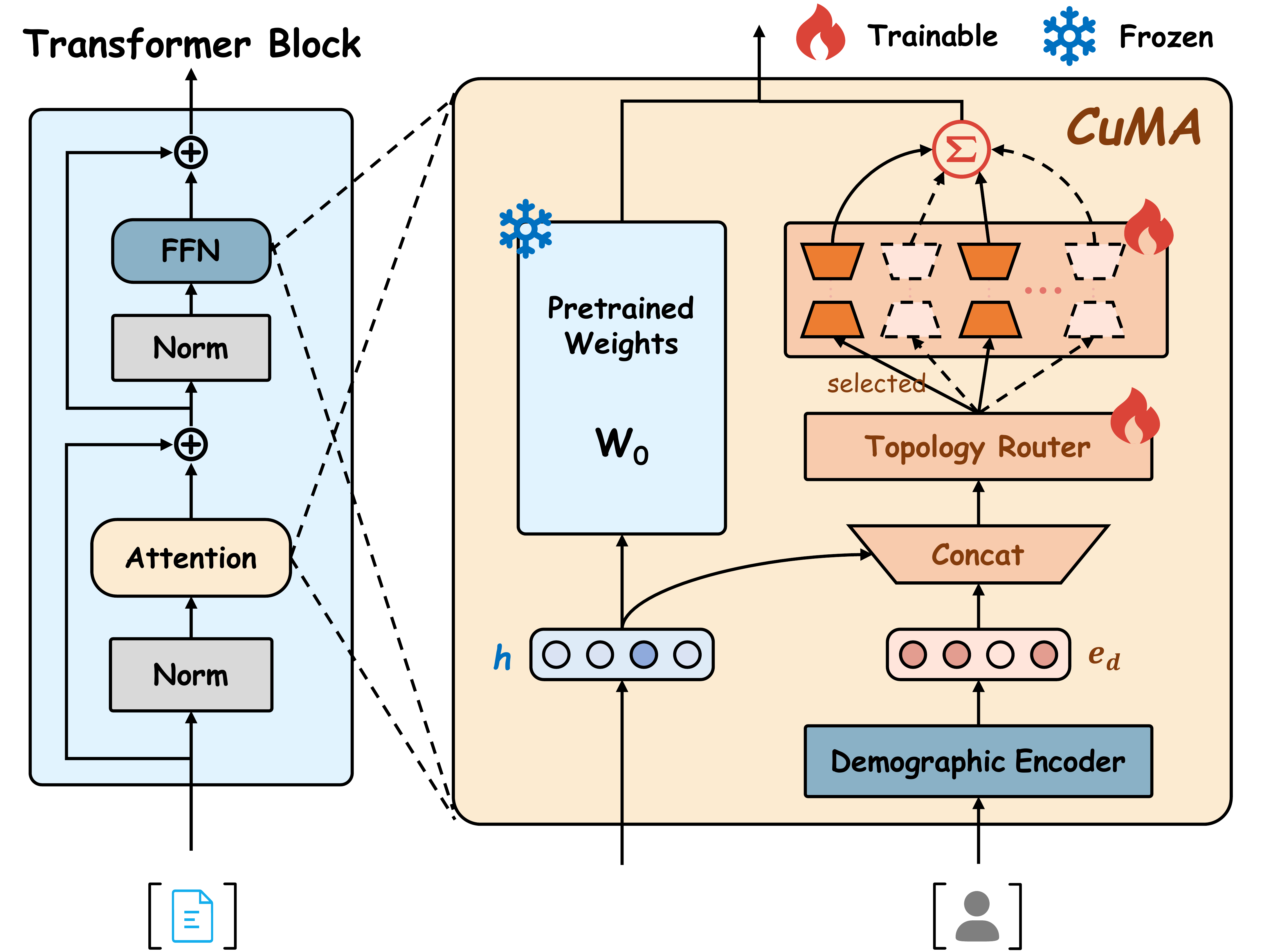}
\caption{\textbf{Architecture of \textsc{CuMA}.} The framework disentangles cultural values by conditioning the routing mechanism on both semantic hidden states and demographic embeddings, effectively isolating gradients into specialized experts.}
\label{fig:model}
\end{figure}

To address Cultural Sparsity and Mean Collapse, we propose \textsc{CuMA} (Figure~\ref{fig:model}). Instead of using a single parameter set for conflicting values, \textsc{CuMA} learns a latent cultural topology and routes inputs to specialized, demographically-aligned adapters. This design disentangles gradient interference and preserves the distinct geometry of pluralistic value distributions.

\subsection{Demographic Encoder}
\label{sec:encoder}

To encode diverse demographic profiles and support generalization, we leverage the geometric priors in pre-trained sentence embedding models. The raw demographic profile $d$ typically consists of structured attributes (e.g., $\{ \texttt{Country: Thailand,\, Religion: Buddhism,} $ \\ $ \texttt{Age: 55} \}$). We first linearize this structured set into a natural language description $t_d$ (e.g., "\textit{A 55-year-old Buddhist resident of Thailand}"). We then map $t_d$ to a dense vector representation $e_d \in \mathbb{R}^m$ via a frozen pre-trained embedding model $E(\cdot)$:
\begin{equation}
e_d = E(t_d)
\end{equation}

By utilizing the frozen embedding space, we preserve the semantic topology from pre-training. Within this space, culturally related groups naturally cluster based on shared traits like geography or religion. This stable structure provides robust signals for the router to measure similarity, enabling generalization to unseen demographic groups.

\subsection{Router as Topology Learner}
\label{sec:router}

The router serves as the core topological mapper. Unlike standard MoE routers that dispatch tokens based solely on internal hidden states (semantic content), our router learns the latent cultural topology by conditioning on the joint interaction between the semantic context and the demographic profile.

For a given layer input $h \in \mathbb{R}^H$ and demographic embedding $e_d$, the router computes the routing logits $s \in \mathbb{R}^N$:
\begin{equation}
s = W_r \cdot [h \oplus e_d]
\end{equation}
where $\oplus$ denotes concatenation and $W_r$ is the learnable routing matrix. This joint representation allows the router to disentangle what is being asked ($h$) from who is asking ($e_d$).

To enforce the conditional capacity separation, we activate only the Top-$k$ experts. The sparse gating weights $g$ are computed via a softmax normalization over the selected experts:
\begin{equation}
g_i = \frac{\exp(s_i) \cdot \mathds{1}[i \in \text{Top-}k(s)]}{\sum_{j=1}^N \exp(s_j) \cdot \mathds{1}[j \in \text{Top-}k(s)]}
\label{eq:router}
\end{equation}

Guided by the latent cultural topology learned in $W_r$, the router directs divergent cultural modes to distinct expert subsets, thereby structurally isolating conflicting gradients and preventing interference.

\subsection{Mixture of Cultural Adapters}
\label{sec:moe}

To enable fine-grained adaptation while preserving general reasoning, we freeze the backbone weights $W_0 \in \mathbb{R}^{d_{out} \times d_{in}}$ and adopt a modular parameter-efficient strategy. We instantiate the expert pool using Low-Rank Adaptation (LoRA)~\cite{hu2021loralowrankadaptationlarge}, chosen for its proven stability and efficiency in large-scale fine-tuning tasks.

Formally, a standard LoRA module modulates the frozen weights by learning a low-rank update $\Delta W = BA$, where $B \in \mathbb{R}^{d_{out} \times r}$ and $A \in \mathbb{R}^{r \times d_{in}}$ are trainable matrices with rank $r \ll \min(d_{in}, d_{out})$. We extend this formulation to a Mixture of LoRA Experts. We initialize $N$ distinct expert modules, denoted as $\{ (A_i, B_i) \}_{i=1}^N$. Guided by the sparse routing weights $g$ (Eq.~\ref{eq:router}), the forward pass for a hidden state $h$ becomes:
\begin{equation}
h' = W_0 h + \sum_{i=1}^N g_i \cdot \underbrace{(B_i A_i h)}_{\text{Expert } i}
\end{equation}

\textsc{CuMA} constructs a demographic-aware update $\Delta W(d) = \sum g_i(d) B_i A_i$. This ensures that conflicting cultural values are processed by separate parameter combinations, directly preventing the gradient interference that causes mean collapse.

\subsection{Optimization Objectives}
\label{sec:training}

\textsc{CuMA} adopts a flexible optimization strategy designed to accommodate varying data granularities. The training process establishes foundational alignment via Conditional Supervised Fine-Tuning (SFT), which can be further refined through Conditional Preference Optimization when preference annotations or group-based rewards are available. The complete training procedure, detailing the curriculum transition and objective selection, is summarized in Appendix~\ref{sec:appendix_training}.

Accordingly, the generalized objective function is a weighted combination of the active task loss and an auxiliary load-balancing regularization:
\begin{equation}
\mathcal{L} = \mathcal{L}_{\text{task}} + \lambda_{\text{lb}}\mathcal{L}_{\text{lb}}
\end{equation}
where $\mathcal{L}_{\text{task}}$ corresponds to either the SFT, DPO, or GRPO objective depending on the training stage. We provide the detailed formulations for each objective component and the full training algorithm in Appendix~\ref{sec:appendix_training}.

\section{Experimental Setup}
\label{sec:setup}

Our experiments are designed to investigate the nature of cultural sparsity and evaluate the efficacy of conditional capacity separation. Specifically, we aim to answer the following three research questions (\textbf{RQs}):

\begin{itemize}
    \item \textbf{RQ1:} Can \textsc{CuMA} achieve superior cultural alignment compared to dense baselines across diverse benchmarks, and how does it perform under varying data scales?
    \item \textbf{RQ2:} How does \textsc{CuMA} mitigate \textit{mean collapse} to avoid the generic, uncertain response patterns of dense models, and to what extent does it preserve the intrinsic diversity of cultural value distributions?
    \item \textbf{RQ3:} Does the demographic-aware router successfully capture the latent cultural topology and enable generalization to unseen demographic groups?
\end{itemize}

\subsection{Datasets and Metrics}
We evaluate \textsc{CuMA} on three benchmarks using a 10:1 train/test split; see Appendix~\ref{sec:appendix_stats} for detailed statistics.

\paragraph{WorldValuesBench (WVB):} Derived from the World Values Survey, this benchmark evaluates value prediction across distinct cultural regions~\cite{zhao-etal-2024-worldvaluesbench}. Given a demographic profile, the model predicts the value stance on a multiple-choice scale. \textbf{Metrics:} We report Accuracy and Macro-F1. Additionally, acknowledging the ordinal nature of Likert-scale responses~\cite{zhao-etal-2024-worldvaluesbench}, we report the Wasserstein-1 Distance (e.g., Earth Mover's Distance (EMD)). This metric quantifies the structural divergence between the model's predicted probabilities and the human value distribution, where a lower distance indicates superior alignment.

\paragraph{Community Alignment (CA):} This dataset \cite{zhang2025cultivatingpluralismalgorithmicmonoculture} captures conflicting preferences of diverse social groups on controversial topics. We evaluate two sub-tasks: preference prediction and response generation. \textbf{Metrics:} We use Accuracy and Macro-F1 for prediction. For generation, we employ a GPT-4o-based\footnote{Model version: \texttt{gpt-4o-2024-11-13}.} judge to compute the pairwise Win-Rate (details in Appendix~\ref{sec:appendix_prompts}). An expert audit on a 100-sample subset with five annotators confirms strong agreement with the automated judge (Cohen's $\kappa = 0.84$; see Appendix~\ref{sec:appendix_human_audit}). We specifically evaluate the preference-optimized models (SFT+DPO and SFT+GRPO) against the base model to assess alignment validity.

\paragraph{PRISM:} PRISM~\cite{NEURIPS2024_be2e1b68} links fine-grained individual profiles to open-ended, multi-turn conversations. \textbf{Metrics:} We report the Win-Rate, adopting the identical evaluation setting as the CA generation task.

\subsection{Baselines}

We compare \textsc{CuMA} against three categories of alignment strategies to isolate performance sources.

\paragraph{Inference-Time Baselines.} These methods steer the base model without parameter updates. We consider: (1) Vanilla Baseline, the unaligned base model representing default pre-training bias; (2) Persona Prompting~\cite{lutz-etal-2025-prompt}, which prepends a demographic-specific system prompt; and (3) Prompt Steering~\cite{miehling2025evaluatingpromptsteerabilitylarge}, employing $k$-shot ($k=3$) demonstrations retrieved from matching demographics to guide the model via analogy (see Appendix~\ref{sec:appendix_prompts}).

\paragraph{Dense Fine-Tuning.} These methods update a single set of global parameters on the combined multicultural dataset. We include: (1) Full Fine-Tuning (FFT), updating 100\% of parameters; (2) P-Tuning v2~\cite{liu2022ptuning}, which optimizes deep prompt vectors; (3) LoRA~\cite{hu2021loralowrankadaptationlarge}, standard Low-Rank Adaptation ($r=64$); and (4) DoRA~\cite{10.5555/3692070.3693369}, which decomposes weights into magnitude and direction components. These methods represent the "one-size-fits-all" parameterization, which we hypothesize is structurally prone to mean collapse.

\paragraph{Sparsely Activated Adapters.} We compare against state-of-the-art MoE-LoRA architectures including (1) MixLoRA~\cite{li2024mixloraenhancinglargelanguage} and (2) HydraLoRA~\cite{NEURIPS2024_123fd8a5}. These models utilize sparse parameter structures but route based solely on semantic hidden states. We include them to verify whether semantic routing alone is sufficient to resolve cultural conflicts, or if explicit demographic conditioning (as in \textsc{CuMA}) is necessary.

\subsection{Implementation Details}

We implement \textsc{CuMA} on two backbones: \texttt{Llama-3.1-8B-Instruct}~\cite{grattafiori2024llama3herdmodels} and \texttt{Qwen3-8B}~\cite{yang2025qwen3technicalreport}. We utilize a frozen \texttt{Qwen3-Embedding-0.6B}~\cite{zhang2025qwen3embeddingadvancingtext} as the demographic encoder. All models are trained on NVIDIA RTX PRO 6000 GPUs. We employ the AdamW optimizer with a cosine decay schedule. For \textsc{CuMA}, we set the number of experts $N=8$ with Top-$k=2$ routing, applying LoRA adapters ($r=8/64$). Detailed hyperparameters and prompt templates are provided in Appendix~\ref{sec:appendix_implementation}.

\section{Results and Analysis}

\begin{table*}[t]
\centering
\resizebox{\textwidth}{!}{%
\setlength{\tabcolsep}{2.5pt} 
\begin{tabular}{l|l|c|ccc|cccc|cc}
\toprule
\multirow{3}{*}{\textbf{Category}} & \multirow{3}{*}{\textbf{Method}} & \multirow{3}{*}{\makecell{\textbf{Trainable} \\ \textbf{Params}}} & \multicolumn{3}{c|}{\textbf{WorldValuesBench (WVB)}} & \multicolumn{4}{c|}{\textbf{Community Alignment (CA)}} & \multicolumn{2}{c}{\textbf{PRISM}} \\ 
\cmidrule(lr){4-6} \cmidrule(lr){7-10} \cmidrule(lr){11-12} 
& & & \multirow{2}{*}{\textbf{Acc} $\uparrow$} & \multirow{2}{*}{\textbf{Macro-F1} $\uparrow$} & \multirow{2}{*}{\textbf{EMD} $\downarrow$} & \multirow{2}{*}{\textbf{Acc} $\uparrow$} & \multirow{2}{*}{\textbf{Macro-F1} $\uparrow$} & \multicolumn{2}{c|}{\textbf{Win-Rate vs Base}} & \multicolumn{2}{c}{\textbf{Win-Rate vs Base}} \\ 
& & & & & & & & \textbf{(DPO)} & \textbf{(GRPO)} & \textbf{(DPO)} & \textbf{(GRPO)} \\ \midrule

\multicolumn{12}{c}{\textsc{\textbf{Backbone: Llama-3.1-8B}}} \\ \midrule
\multirow{3}{*}{\shortstack[l]{\textit{Inference-Time}\\ \textit{Strategies}}} 
  & Vanilla Baseline & 0.00\% & 32.42 & 22.99 & 0.3967 & 26.70 & 20.79 & - & - & - & - \\
  & Persona Prompting & 0.00\% & 37.06 & 23.90 & 0.3105 & 26.10 & 21.57 & 55.5\% & 56.2\% & 55.2\% & 55.8\% \\
  & Prompt Steering (3-shot) & 0.00\% & 27.50 & 11.14 & 0.2507 & 26.80 & 22.74 & 56.8\% & 57.5\% & 56.5\% & 59.2\% \\ \cmidrule{1-12}
\multirow{5}{*}{\shortstack[l]{\textit{Dense}\\ \textit{Fine-Tuning}}} 
  & Full Fine-Tuning (FFT) & 100.0\% & 45.25 & \underline{30.50} & 0.2205 & 45.15 & 32.30 & 63.5\% & 65.2\% & 61.5\% & 63.2\% \\
  & P-Tuning v2 & 0.94\% & 43.80 & 29.10 & 0.2470 & 43.50 & 30.85 & 57.2\% & 58.8\% & 55.5\% & 56.8\% \\
  & LoRA & 0.37\% & 34.30 & 22.37 & 0.2537 & 38.53 & 30.50 & 60.5\% & 62.1\% & 58.8\% & 59.5\% \\
  & DoRA & 0.38\% & 36.50 & 25.10 & 0.2587 & 39.20 & 31.50 & 61.8\% & 63.5\% & 59.5\% & 61.2\% \\ \cmidrule{1-12}
\multirow{4}{*}{\shortstack[l]{\textit{Sparsely}\\ \textit{Activated} \\ \textit{Adapters}}} 
  & MixLoRA & 3.01\% & 45.20 & 29.80 & 0.2440 & 46.80 & 34.60 & 66.5\% & 68.2\% & 64.2\% & 65.8\% \\
  & HydraLoRA & 2.31\% & 46.50 & 29.90 & 0.2350 & 47.90 & 36.20 & \underline{69.8\%} & 69.5\% & 65.5\% & \underline{68.2\%} \\
  & \textsc{CuMA} ($r=8$) & 1.53\% & \underline{48.90} & \underline{30.50} & \underline{0.1903} & \underline{50.12} & \underline{38.50} & 68.5\% & \underline{73.8\%} & \underline{68.8\%} & 67.5\% \\
  & \textsc{CuMA} & 4.15\% & \textbf{50.46} & \textbf{32.50} & \textbf{0.1870} & \textbf{52.45} & \textbf{40.12} & \textbf{72.2\%} & \textbf{74.5\%} & \textbf{71.2\%} & \textbf{73.5\%} \\ \midrule

\multicolumn{12}{c}{\textsc{\textbf{Backbone: Qwen3-8B}}} \\ \midrule
\multirow{3}{*}{\shortstack[l]{\textit{Inference-Time}\\ \textit{Strategies}}} 
  & Vanilla Baseline & 0.00\% & 31.68 & 18.92 & 0.3851 & 31.20 & 17.75 & - & - & - & - \\
  & Persona Prompting & 0.00\% & 34.92 & 21.05 & 0.2864 & 32.80 & 21.00 & 57.1\% & 58.5\% & 56.2\% & 57.0\% \\
  & Prompt Steering (3-shot) & 0.00\% & 28.08 & 12.36 & 0.2299 & 26.00 & 22.19 & 59.5\% & 60.8\% & 58.4\% & 59.5\% \\ \cmidrule{1-12}
\multirow{4}{*}{\shortstack[l]{\textit{Dense}\\ \textit{Fine-Tuning}}} 
  & Full Fine-Tuning (FFT) & 100.0\% & 45.54 & 28.21 & 0.2228 & 49.50 & 36.20 & 66.8\% & 68.5\% & 63.5\% & 65.2\% \\
  & P-Tuning v2 & 0.94\% & 45.04 & 28.17 & 0.2358 & 47.50 & 34.80 & 59.5\% & 61.2\% & 57.5\% & 58.8\% \\
  & LoRA & 0.37\% & 40.06 & 22.02 & 0.2700 & 38.53 & 30.50 & 63.2\% & 65.5\% & 61.5\% & 62.2\% \\
  & DoRA & 0.38\% & 42.78 & 24.73 & 0.2773 & 39.20 & 31.50 & 64.5\% & 66.8\% & 62.8\% & 64.1\% \\ \cmidrule{1-12}
\multirow{4}{*}{\shortstack[l]{\textit{Sparsely}\\ \textit{Activated} \\ \textit{Adapters}}} 
  & MixLoRA & 3.01\% & 43.50 & 26.44 & 0.2904 & 51.50 & 38.80 & 70.5\% & 72.8\% & 67.5\% & 69.2\% \\
  & HydraLoRA & 2.31\% & 45.36 & 28.12 & 0.2793 & 52.80 & 40.20 & 71.5\% & 73.6\% & 68.5\% & 70.4\% \\
  & \textsc{CuMA} ($r=8$) & 1.53\% & \underline{49.02} & \underline{29.70} & \underline{0.1980} & \underline{55.40} & \underline{43.10} & \underline{75.8\%} & \underline{76.5\%} & \underline{73.2\%} & \underline{75.5\%} \\
  & \textsc{CuMA} & 4.15\% & \textbf{50.64} & \textbf{31.50} & \textbf{0.1876} & \textbf{57.20} & \textbf{44.80} & \textbf{77.5\%} & \textbf{78.2\%} & \textbf{74.5\%} & \textbf{76.8\%} \\ \bottomrule
\end{tabular}%
}
\caption{\textbf{Main Results on Cultural Alignment Benchmarks.} Comparison of \textsc{CuMA} against static, dense, and sparse baselines across two backbones: \textbf{Llama-3.1-8B} and \textbf{Qwen3-8B}. \textbf{Trainable Params} denotes the exact percentage of trainable parameters relative to the base model. Standard LoRA, DoRA, and \textsc{CuMA} imply rank $r=64$ unless specified otherwise ($r=8$). For Win-Rates, we report results after DPO and GRPO stages respectively. \textbf{Bold} indicates the best performance, and \underline{underline} indicates the second best performance.}
\label{tab:main_results}
\vspace{-2mm}
\end{table*}

In this section, we present empirical findings addressing our research questions. We first evaluate \textsc{CuMA}'s overall efficacy against baselines (\textbf{RQ1}), then analyze its ability to mitigate mean collapse and preserve diversity (\textbf{RQ2}). We further investigate the learned latent topology and its generalization capabilities (\textbf{RQ3}), concluding with ablation studies on key architectural components.

\subsection{Overall Alignment Performance}

Table~\ref{tab:main_results} summarizes results across three benchmarks, showing consistent trends for both Llama-3.1-8B and Qwen3-8B.

\paragraph{Structural Limitations of Dense Models.}
Dense methods (FFT, LoRA, DoRA) show a distinct performance ceiling. On Llama-3.1 WVB, even Full Fine-Tuning (45.25\% Acc) lags significantly behind \textsc{CuMA} (50.46\% Acc). This saturation indicates a structural bottleneck: the "one-size-fits-all" parameterization suffers from gradient interference when optimizing for conflicting values, forcing convergence towards an averaged solution rather than distinct cultural modes.

\paragraph{Efficiency of Demographic Conditioning.}
\textsc{CuMA} proves that alignment depends on routing precision, not just parameter scale. The low-rank variant ($r{=}8$, 1.53\% params) consistently outperforms the larger HydraLoRA (2.31\% params), e.g., +2.4\% Acc on Llama-3.1 WVB. This confirms that conditioning routing on demographic topology allocates capacity more effectively than semantic-only MoEs, achieving superior results with fewer parameters.

\paragraph{Mitigating Semantic Stereotyping.}
A critical divergence appears between Accuracy and EMD in baselines. Semantic sparse methods (MixLoRA, HydraLoRA) achieve competitive Accuracy but suffer high EMD (e.g., 0.28 vs. 0.19 for \textsc{CuMA} on Qwen3). This "High-Accuracy, High-EMD" pattern suggests "stereotyping": models predict the mode based on semantics but miss the nuanced probability spread. \textsc{CuMA}'s superior EMD indicates it successfully models the diverse shape of human value distributions rather than memorizing stereotypes.

\paragraph{Holistic Alignment across Modalities.}
This distributional fidelity translates to robust generation. With DPO/GRPO, \textsc{CuMA} achieves dominant Win-Rates on CA (78.2\%) and PRISM (76.8\%) with Qwen3, surpassing dense baselines ($\approx$ 65\%). This verifies \textsc{CuMA}'s ability to map latent values into coherent, culturally aligned responses.

\subsection{Verification of Mean Collapse}
\label{sec:collapse_analysis}

To address \textbf{RQ2}, we employ Prediction Entropy (WVB) and Distinct-2 scores (CA-generation/PRISM) to diagnose mean collapse.

\begin{figure}[t]\centering\includegraphics[width=0.95\linewidth]{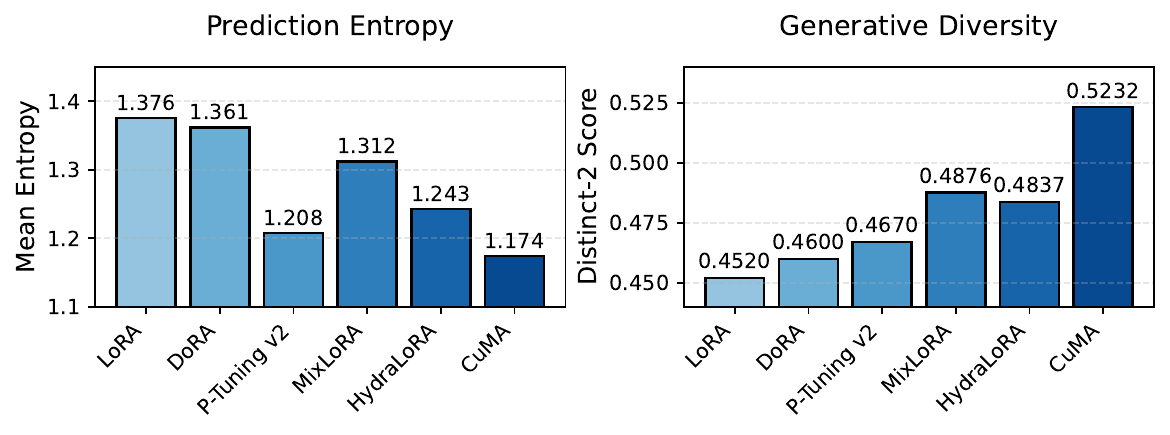}\caption{\textbf{Quantitative Verification of Mean Collapse.} \textbf{(Left)} Dense baselines (e.g., LoRA, DoRA) exhibit high prediction entropy ($H \approx$ 1.38), indicating probability mass dispersion typical of mean collapse. \textsc{CuMA} significantly reduces uncertainty ($H \approx$ 1.17). \textbf{(Right)} In open-ended generation, \textsc{CuMA} achieves the highest Distinct-2 score, confirming that it avoids repetitive, generic templates by accessing specialized cultural vocabularies.}
\label{fig:collapse}
\end{figure}

As shown in Figure~\ref{fig:collapse}, dense models exhibit high entropy ($H_{\text{mean}} \approx$ 1.38), reflecting the "diluted middle" behavior predicted in Appendix~\ref{sec:appendix_variance}. In the context of next-token prediction, high entropy indicates a flat, indecisive distribution where the model hedges across options rather than committing to a culturally specific response, which is a symptom of variance inflation from merging conflicting modes. \textsc{CuMA} reduces entropy to 1.17, indicating sharper, more decisive alignment for each demographic profile. Crucially, this per-profile decisiveness does not sacrifice cross-profile diversity: \textsc{CuMA} achieves a Distinct-2 score of 0.52, outperforming dense baselines ($\approx$ 0.45), confirming that it produces distinct, culturally resonant outputs rather than repetitive generic templates.

\subsection{Latent Cultural Topology and Generalization}
\label{sec:topology}

\begin{figure}[t]
\centering
\includegraphics[width=0.90\linewidth]{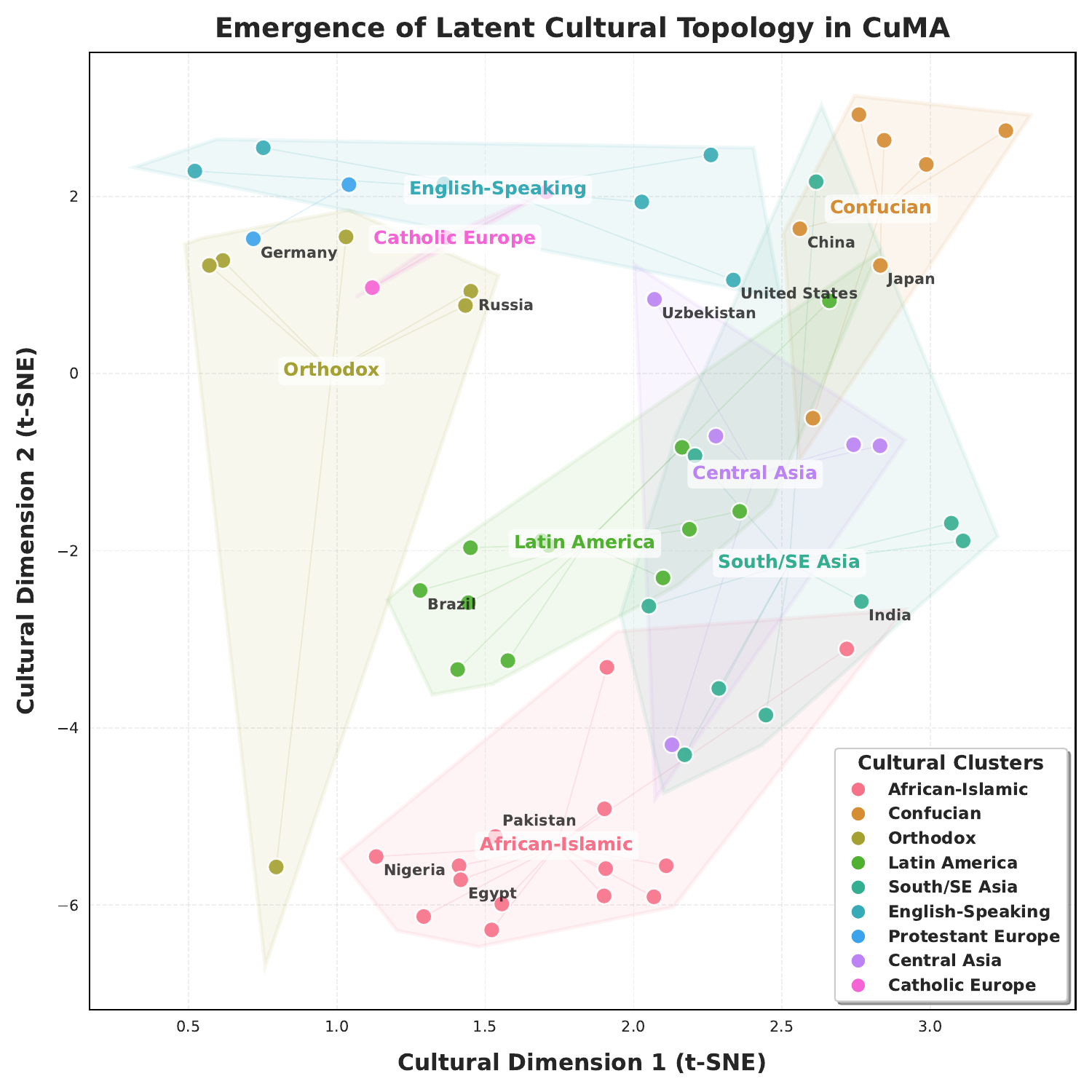}
\caption{\textbf{Emergence of Latent Cultural Topology.} t-SNE projection of expert activation patterns across 65 nations. Without explicit supervision, the router spontaneously organizes demographic profiles into coherent clusters that align with sociological frameworks (e.g., the \textit{African-Islamic} and \textit{Confucian} spheres). This geometric structure facilitates zero-shot generalization by routing unseen demographic profiles to experts trained on culturally proximate groups. Details on the visualization protocol are provided in Appendix~\ref{sec:appendix_analysis}.}
\label{fig:topology}
\end{figure}

To address \textbf{RQ3}, we investigate the learned geometric representation and its generalization potential.

\paragraph{Visualizing the Latent Topology.}
Figure~\ref{fig:topology} visualizes expert activation patterns for 65 countries via t-SNE. The router spontaneously organizes demographics into clusters aligning with sociological frameworks (e.g., Inglehart--Welzel~\cite{Inglehart_Welzel_2005}), such as the African-Islamic bloc and Confucian sphere. This confirms the construction of a latent cultural topology, where groups with shared value affinities share model capacity without explicit supervision.

\paragraph{Quantitative Verification: Zero-Shot Transfer.}
We validate generalization by evaluating on held-out demographic profiles (Table~\ref{tab:generalization_full}). Despite lacking supervision for these specific profiles, \textsc{CuMA} exhibits robust topological transfer. Under the unweighted macro-average across clusters, accuracy drops by only 2.12\% and EMD increases by +0.0244. The English-Speaking cluster shows the smallest drop (-1.67\%), benefiting from richer pre-training representation and higher sociological compactness on the Inglehart-Welzel cultural map~\cite{Inglehart_Welzel_2005}. Even the more dispersed African-Islamic cluster degrades only marginally (-2.36\%).

To contextualize these results, Table~\ref{tab:zero_shot_baselines} compares \textsc{CuMA}'s zero-shot performance against all baselines under the same held-out demographic profile protocol. For consistency with the main results table, this comparison uses sample-weighted averages. Under this aggregation, \textsc{CuMA} achieves 48.10\% zero-shot accuracy, significantly outperforming the strongest zero-shot baseline (HydraLoRA, 36.2\%) and even surpassing the full-supervision accuracy of several dense baselines (e.g., FFT: 45.54\%; LoRA: 40.06\%).

\begin{table}[t]
\centering
\resizebox{\columnwidth}{!}{%
\setlength{\tabcolsep}{3.5pt}
\begin{tabular}{l|cc|cc|cc}
\toprule
\multirow{2}{*}{\textbf{Cultural Cluster}} & \multicolumn{2}{c|}{\textbf{Full Sup.}} & \multicolumn{2}{c|}{\textbf{Zero-Shot}} & \multicolumn{2}{c}{\textbf{Gap ($\Delta$)}} \\ 
 & \textbf{Acc} $\uparrow$ & \textbf{EMD} $\downarrow$ & \textbf{Acc} $\uparrow$ & \textbf{EMD} $\downarrow$ & \textbf{Acc} & \textbf{EMD} \\ \midrule
African-Islamic & 53.81 & 0.2244 & 51.45 & 0.2510 & -2.36 & +0.0266 \\
Catholic Europe & 47.98 & 0.2110 & 45.82 & 0.2350 & -2.16 & +0.0240 \\
Central Asia & 49.72 & 0.2808 & 47.55 & 0.3090 & -2.17 & +0.0282 \\
Confucian & 48.71 & 0.2387 & 46.60 & 0.2640 & -2.11 & +0.0253 \\
English-Speaking & 50.82 & 0.1970 & 49.15 & 0.2150 & -1.67 & +0.0180 \\
Latin America & 49.77 & 0.2568 & 47.65 & 0.2810 & -2.12 & +0.0242 \\
Orthodox & 49.87 & 0.2368 & 47.90 & 0.2610 & -1.97 & +0.0242 \\
Protestant Europe & 50.57 & 0.2182 & 48.35 & 0.2410 & -2.22 & +0.0228 \\
South/SE Asia & 50.39 & 0.2316 & 48.10 & 0.2580 & -2.29 & +0.0264 \\ \midrule
Macro Avg. & 50.18 & 0.2328 & 48.06 & 0.2572 & -2.12 & +0.0244 \\ \bottomrule
\end{tabular}%
}
\caption{\textbf{Zero-Shot Cross-Cultural Generalization.} Results of the zero-shot generalization experiment across 9 cultural clusters. \textbf{Full Sup.} indicates standard training, while \textbf{Zero-Shot} evaluates performance on held-out demographic profiles excluded during training. The final row reports the unweighted macro-average over cultural clusters. \textbf{Gap ($\Delta$)} denotes the performance difference between Full Supervision and Zero-Shot. The minimal degradation (Macro Avg. $\Delta_{\text{Acc}} \approx$ -2.1\%) confirms that \textsc{CuMA} effectively generalizes to unseen cultures by leveraging the latent topology. See Appendix~\ref{sec:appendix_analysis} for experimental details.}
\label{tab:generalization_full}
\end{table}

\begin{table}[t]
\centering
\resizebox{\columnwidth}{!}{%
\setlength{\tabcolsep}{3pt}
\begin{tabular}{l|cc|cc|cc}
\toprule
\multirow{2}{*}{\textbf{Method}} & \multicolumn{2}{c|}{\textbf{Full Sup.}} & \multicolumn{2}{c|}{\textbf{Zero-Shot}} & \multicolumn{2}{c}{\textbf{Gap ($\Delta$)}} \\
 & \textbf{Acc} $\uparrow$ & \textbf{EMD} $\downarrow$ & \textbf{Acc} $\uparrow$ & \textbf{EMD} $\downarrow$ & \textbf{Acc} & \textbf{EMD} \\ \midrule
LoRA & 40.06 & 0.2700 & 32.40 & 0.3250 & -7.66 & +0.0550  \\
DoRA & 42.78 & 0.2773 & 33.12 & 0.3180 & -9.66 & +0.0407 \\
P-Tuning v2 & 45.04 & 0.2358 & 30.50 & 0.3320 & -14.54 & +0.0962 \\
MixLoRA & 43.50 & 0.2904 & 35.80 & 0.2820 & -7.70 & -0.0084 \\
HydraLoRA & 45.36 & 0.2793 & 36.24 & 0.2750 & -9.12 & -0.0043 \\ \midrule
\textsc{CuMA} (ours) & 50.64 & 0.1876 & 48.10 & 0.2100 & -2.54 & +0.0224 \\ \bottomrule
\end{tabular}%
}
\caption{\textbf{Zero-Shot Generalization vs. Baselines on WVB (Qwen3-8B).} All methods are evaluated under the same held-out demographic profile protocol. The table reports sample-weighted averages following the aggregation protocol of Table~\ref{tab:main_results}. \textsc{CuMA}'s zero-shot accuracy (48.10\%) surpasses the full-supervision accuracy of most dense baselines, demonstrating the effectiveness of the learned latent topology.}
\label{tab:zero_shot_baselines}
\end{table}

\subsection{Ablation Studies}
\label{sec:ablation}

We validate \textsc{CuMA}'s components on Qwen3-8B by ablating the demographic routing branch ($e_d$), semantic routing ($h$), and auxiliary load balancing loss ($\mathcal{L}_{\text{aux}}$). Table~\ref{tab:ablation} summarizes the results.

\begin{table}[h]
\centering
\resizebox{\columnwidth}{!}
{
\begin{tabular}{l|ccc}
\toprule
\textbf{Method} & \textbf{Acc} $\uparrow$ & \textbf{Macro-F1} $\uparrow$ & \textbf{EMD} $\downarrow$ \\
\midrule
\textbf{\textsc{CuMA} (Full)} & \textbf{50.64} & \textbf{31.50} & \textbf{0.1876} \\
\midrule
w/o Demographic Routing & 47.08 & 29.98 & 0.1965 \\
w/o Demo. \& Bal. Loss & 45.26 & 27.49 & 0.2657 \\
w/o Semantic Routing & 44.26 & 22.99 & 0.3060 \\
Full Cancellation & 32.15 & 19.25 & 0.3518 \\ 
\bottomrule
\end{tabular}
}
\caption{\textbf{Ablation Studies on WVB (Qwen3-8B).} We evaluate the impact of removing demographic routing, semantic routing, and the load balancing loss. \textit{"w/o Demo. \& Bal. Loss"} represents a naive semantic MoE without auxiliary loss. \textit{"Full Cancellation"} denotes the removal of all routing mechanisms and demographic prompts.}
\label{tab:ablation}
\end{table}

Results demonstrate the synergy between semantic and demographic signals. Removing demographic routing ($w/o$ Demo.) acts as a standard semantic MoE, dropping accuracy by 3.56\%. This confirms that resolving cultural conflict requires explicit demographic conditioning. Conversely, in the $w/o$ Semantic Routing setting, we replace the demographic-specific prompt with a generic instruction ("\textit{You are a helpful assistant that answers survey questions honestly}"), forcing reliance solely on demographic embeddings. This causes a larger accuracy drop (-6.38\%), yet still significantly outperforms the random baseline (Full Cancellation), proving that the router successfully captures latent value priors solely from the demographic topology. Finally, removing the auxiliary balancing loss ($w/o$ Demo. \& Bal. Loss) spikes EMD (0.1876 $\to$ 0.2657), indicating that structural regularization is critical for preventing mode collapse and ensuring effective expert utilization. We further analyze the impact of the routing strategy (e.g., Soft vs. Top-k Routing) in Appendix~\ref{sec:appendix_routing}, finding that strict capacity separation (Top-k) is essential for resolving cultural interference.

\section{Related Work}
\label{sec:related_work}

Existing alignment paradigms typically prioritize universal attributes~\cite{ouyang2022training,NEURIPS2023_a85b405e}, often leading to "Algorithmic Monoculture"~\cite{zhang2025cultivatingpluralismalgorithmicmonoculture}. While recent pluralistic alignment methods~\cite{NEURIPS2024_9a16935b,xu-etal-2025-self,NEURIPS2024_be2e1b68,wang-etal-2024-cdeval} attempt to incorporate diverse values, they largely rely on dense parameterizations. Even when utilizing parameter-efficient variations such as LoRA~\cite{hu2021loralowrankadaptationlarge} and DoRA~\cite{10.5555/3692070.3693369}, these methods remain fundamentally "dense" by sharing a unified weight space, which we argue renders them structurally vulnerable to gradient interference and mean collapse.

To address this, we draw upon Mixture-of-Experts (MoE) architectures~\cite{shazeer2017outrageously}. Unlike recent PEFT-MoE approaches~\cite{li2024mixloraenhancinglargelanguage, NEURIPS2024_123fd8a5} that rely on semantic or task-specific routing to enhance multi-task competence, \textsc{CuMA} re-purposes MoE for conditional capacity separation. By conditioning routing on demographic topology, we isolate conflicting cultural gradients, preventing the homogenization of distinct cultural values. A comprehensive review of related work is provided in Appendix~\ref{sec:appendix_related_work}.

\section{Conclusion}

We introduced \textsc{CuMA}, a framework that reformulates cultural alignment as a conditional capacity separation problem. By using demographic-aware routing, \textsc{CuMA} learns a \textit{Latent Cultural Topology} to disentangle conflicting gradients and resolve \textit{Mean Collapse}. Results across three benchmarks show significant gains: \textsc{CuMA} reduces distributional divergence (EMD) to 0.1876 and outperforms dense baselines by over 5\% in accuracy. It also achieves dominant Win-Rates on Community Alignment (78.2\%) and PRISM (76.8\%). These findings suggest that respecting the sparsity of cultural values is key to building truly pluralistic LLMs.

\section*{Limitations}

While \textsc{CuMA} demonstrates significant improvements in cultural alignment, several limitations remain. First, the framework relies on explicit demographic profiles to guide the routing mechanism. In real-world scenarios, such information may be incomplete, inaccurate, or unavailable due to privacy constraints. Second, our experiments utilized a fixed number of experts ($N=8$). While this capacity proved sufficient for the benchmarks studied, capturing the full complexity of global cultural diversity may require more granular expert pools or hierarchical routing structures. Third, although \textsc{CuMA} generalizes well to unseen demographic groups, its performance is still bounded by the coverage and potential biases of the underlying training datasets (WVB, CA, and PRISM). Finally, the MoE-based architecture introduces modest inference overhead (see Appendix~\ref{sec:appendix_efficiency} for detailed profiling). Future work will explore implicit demographic inference and dynamic expert allocation to further enhance the flexibility of pluralistic alignment.

\section*{Acknowledgments}

Yuheng Jia was supported by the National Natural Science Foundation of China (Grant U24A20322, Grant 62576094). Shu Su was supported by the National Natural Science Foundation of China (Grant 72371072). This research work was also supported by the Big Data Computing Center of Southeast University.


\bibliography{custom}

\appendix



\section{Extended Related Work}
\label{sec:appendix_related_work}

\subsection{From Universal to Pluralistic Alignment}
The dominant paradigm in LLM alignment has prioritized universal attributes such as helpfulness and safety, typically optimized via Reinforcement Learning from Human Feedback (RLHF) or Direct Preference Optimization (DPO)~\cite{ouyang2022training, NEURIPS2023_a85b405e}. While effective for objective tasks, this "one-size-fits-all" approach fails to encompass the normative diversity of global users, often collapsing into a specific Western-centric value system, a phenomenon termed "Algorithmic Monoculture"~\cite{zhang2025cultivatingpluralismalgorithmicmonoculture}.

In response, recent research has pivoted towards \textit{pluralistic alignment}. This transition is supported by emerging evaluation frameworks: CDEval~\cite{wang-etal-2024-cdeval} and NaVAB~\cite{ju2025benchmarking} assess cultural knowledge and bias, while PRISM~\cite{NEURIPS2024_be2e1b68} links fine-grained sociodemographics to interactive preferences. On the methodological front, approaches like CultureLLM~\cite{NEURIPS2024_9a16935b} utilize semantic data augmentation, and CultureSPA~\cite{xu-etal-2025-self} employs contrastive learning to distinguish cultural norms. Others have explored personalization, predicting individual value judgments from historical context~\cite{jiang-etal-2025-language}.

However, a critical structural gap remains. Most existing methods treat cultural alignment as a data scale or prompting problem, attempting to inject pluralistic cultural values into a dense model. They overlook the inherent conflict arising from this multiplicity: since these values are often mutually exclusive, forcing a single set of parameters to represent them leads to gradient interference. Without structural separation, these methods remain vulnerable to mean collapse.

\subsection{Parameter-Efficient MoE for Value Disentanglement}
To address parameter interference, Mixture-of-Experts (MoE) architectures have seen renewed interest, particularly when combined with Parameter-Efficient Fine-Tuning (PEFT). LoRA~\cite{hu2021loralowrankadaptationlarge} provides a lightweight adaptation mechanism, while MoE scales capacity via conditional computation~\cite{shazeer2017outrageously}.

Recent innovations like MixLoRA~\cite{li2024mixloraenhancinglargelanguage} and HydraLoRA~\cite{NEURIPS2024_123fd8a5} integrate these paradigms, composing multiple LoRA adapters to handle diverse downstream tasks. While structurally similar to our approach, these methods employ experts as functional components to maximize multi-task competence. In contrast, \textsc{CuMA} re-purposes the MoE framework for structural value separation. We conceptualize experts not merely as skill specialists, but as culturally specialized parameter spaces that isolate conflicting cultural gradients. By conditioning routing on demographic topology rather than just semantic complexity, \textsc{CuMA} prevents the homogenization of distinct cultural perspectives, mitigating a key limitation in pluralistic alignment.

\section{Derivations of Mean Collapse and Its Resolution}
\label{sec:appendix_derivations}

In Section 2.3, we qualitatively defined \textit{Mean Collapse} as the convergence of a dense model to the statistical average of conflicting modes. In this appendix, we provide the rigorous mathematical derivation of this phenomenon under \textit{Cultural Sparsity} and theoretically demonstrate how \textsc{CuMA}'s conditional routing resolves this structural limitation.

\subsection{Setup: The Mixture Problem}
\label{sec:appendix_setup}

Let the true distribution of human values $P_{\text{data}}(y)$ be a mixture of $K$ distinct cultural modes. For analytical tractability, we approximate these modes as Gaussians. Consider a simplified case with two conflicting groups ($K=2$) with proportions $\pi_1, \pi_2$ (where $\pi_1 + \pi_2 = 1$):
\begin{equation}
P_{\text{data}}(y) = \pi_1 \mathcal{N}(y; \mu_1, \Sigma) + \pi_2 \mathcal{N}(y; \mu_2, \Sigma)
\end{equation}
where $\mu_1, \mu_2$ represent conflicting value centers in the feature space.

A standard dense model $P_\theta(y|x, d)$ utilizes a monolithic parameter set $\theta$ for all groups. Consequently, conflicting gradients from diverse groups interfere within the shared capacity. To analyze this structural tendency, we approximate the dense estimator as a single Gaussian $\mathcal{N}(y; \mu_\theta, \Sigma_\theta)$ optimized via the Forward Kullback-Leibler (KL) divergence:
\begin{equation}
\begin{aligned}
\min_\theta & D_{\text{KL}}(P_{\text{data}} \| P_\theta) \\
&\iff \min_\theta \mathbb{E}_{y \sim P_{\text{data}}} [ -\log P_\theta(y) ]
\end{aligned}
\end{equation}

\subsection{Optimization Dynamics of Dense Models}
\label{sec:appendix_optimization}

We determine the optimal location parameter $\mu_\theta^*$ by minimizing $\mathcal{J}(\mu_\theta) = \mathbb{E}_{y \sim P_{\text{data}}} [ -\log P_\theta(y) ]$. Substituting the Gaussian log-likelihood (ignoring constant terms):
\begin{equation}
\begin{aligned}
\mathcal{J}(\mu_\theta) = & \int P_{\text{data}}(y) \Bigl[ \frac{1}{2} (y - \mu_\theta)^\top \\
& \Sigma_\theta^{-1} (y - \mu_\theta) \Bigr] dy
\end{aligned}
\end{equation}

Taking the gradient with respect to $\mu_\theta$ and setting it to zero:
\begin{equation}
\nabla_{\mu_\theta} \mathcal{J} = -\Sigma_\theta^{-1} \left( \int P_{\text{data}}(y)\, y \, dy - \mu_\theta \right) = 0
\end{equation}

Since $\Sigma_\theta^{-1}$ is positive definite and $\int P_{\text{data}}(y) \, dy = 1$, we obtain:
\begin{equation}
\mu_\theta^* = \int P_{\text{data}}(y) y \, dy = \mathbb{E}_{P_{\text{data}}}[y]
\end{equation}

Expanding the expectation over the mixture components:
\begin{equation}
\mu_\theta^* = \pi_1 \mu_1 + \pi_2 \mu_2
\end{equation}
\hfill $\square$

This proves that the dense model strictly converges to the linearly weighted average of the modes. Regardless of the semantic distance between cultural groups, the single set of parameters is mathematically forced to the geometric center.

\subsection{Geometric Consequences under Cultural Sparsity}
\label{sec:appendix_geometric}

We now analyze the implications of this convergence when the data satisfies the \textit{Cultural Sparsity} condition (large separation $\delta = \|\mu_1 - \mu_2\|$).

\paragraph{1. Probability Density Gap.}
\label{sec:appendix_density_gap}
Assume a symmetric conflict where $\pi_1 = \pi_2 = 0.5$ and $\Sigma = I$. The optimal dense mean lies at $\mu_\theta^* = (\mu_1 + \mu_2)/2$. The distance from this collapsed mean to a true mode is $\|\mu_\theta^* - \mu_1\| = \delta/2$.

The true probability density at the collapsed mean is:
\begin{equation}
\begin{split}
P_{\text{data}}(\mu_\theta^*) &= \frac{1}{2} \mathcal{N}(\mu_\theta^*; \mu_1, I) + \frac{1}{2} \mathcal{N}(\mu_\theta^*; \mu_2, I) \\
&\propto \exp\left( -\frac{1}{2} \left\| \frac{\delta}{2} \right\|^2 \right) = \exp\left( -\frac{\delta^2}{8} \right)
\end{split}
\end{equation}

In contrast, the density at a true mode (e.g., $\mu_1$) is dominated by the first component:
\begin{equation}
P_{\text{data}}(\mu_1) \approx \frac{1}{2} \mathcal{N}(\mu_1; \mu_1, I) \propto \frac{1}{2} \exp(0) = \frac{1}{2}
\end{equation}

The likelihood ratio of the "average" response versus a culturally specific response decays exponentially:
\begin{equation}
\frac{P_{\text{data}}(\mu_\theta^*)}{P_{\text{data}}(\mu_1)} \approx 2 \exp\left( -\frac{\delta^2}{8} \right)
\end{equation}

Under Cultural Sparsity (Eq. 2), where $\delta$ significantly exceeds the ambient dimension ($\delta^2 \gg m$), this ratio vanishes. The dense model effectively hallucinates a "safe middle" that corresponds to a low-density void in the cultural manifold.

\paragraph{2. Variance Inflation.}
\label{sec:appendix_variance}
Mean collapse also implies a loss of precision. By the law of total variance, the optimal covariance $\Sigma_\theta^*$ for the dense model decomposes into two terms:
\begin{equation}
\label{eq:dense_variance}
\begin{split}
\Sigma_\theta^* &= \text{Var}_{P_{\text{data}}}[y] \\
&= \sum_k \pi_k \Sigma_k + \sum_k \pi_k (\mu_k - \mu_\theta^*)(\mu_k - \mu_\theta^*)^\top
\end{split}
\end{equation}

The second term scales quadratically with $\delta$. This forces the dense model to expand its probability mass to span distant modes, exhibiting Maximum Entropy behavior, generating generic, non-committal responses.

\subsection{Resolution via Conditional Routing}
\label{sec:appendix_resolution}

\textsc{CuMA} resolves this dilemma by introducing a conditioning variable $d$ (demographics). The routing mechanism $g(d)$ partitions the parameter space, modeling the conditional density:
\begin{equation}
P_{\text{CuMA}}(y|x, d) \approx \sum_{i} g_i(d) \mathcal{N}(y; \mu_i, \Sigma_i)
\end{equation}

If the router successfully learns the topology (i.e., $g_k(d) \approx \mathds{1}[d \in \text{Group}_k]$), the objective function decomposes into separate objectives for each expert. This allows each expert to converge to the true mode $\mu_k$ and intrinsic covariance $\Sigma_k$ of its respective group.

Crucially, the resulting variance for \textsc{CuMA} becomes:
\begin{equation}
\Sigma_{\text{CuMA}}^* \approx \sum_k \pi_k \Sigma_k
\end{equation}

Comparing this to Eq. \ref{eq:dense_variance}, \textsc{CuMA} explicitly eliminates the structural uncertainty term ($\sum \pi_k (\mu_k - \mu_\theta^*)^2$). By removing this variance inflation, \textsc{CuMA} avoids the exponential density decay and maintains high fidelity to distinct cultural modes.

\section{Detailed Optimization Objectives}
\label{sec:appendix_training}

In this section, we provide the detailed formulations for the optimization objectives. The complete training procedure is summarized in Algorithm~\ref{alg:training}.

\paragraph{1. Conditional SFT.}
For standard instruction following and knowledge injection, we minimize the negative log-likelihood conditioned on the demographic profile $d$:
\begin{equation}
\mathcal{L}_{\text{SFT}}(\theta) = -\mathbb{E}_{(x, y, d) \sim \mathcal{D}_{\text{SFT}}} \left[ \log P_\theta(y \mid x, d) \right]
\end{equation}

\paragraph{2. Conditional Preference Optimization.}
To sharpen the decision boundaries between cultural modes and explicitly penalize mean collapse, we align the model with human preferences. Depending on the available data format, we employ one of the following objectives:

\textbf{Option A: Conditional DPO.} When pairwise preference data $(y_w, y_l)$ is available, we apply Direct Preference Optimization (DPO). Our objective contrasts a chosen response $y_w$ against a rejected response $y_l$ under the \textit{same} demographic profile $d$:
\begin{equation}
\begin{split}
\mathcal{L}_{\text{DPO}}(\theta) = -\mathbb{E}\Big[ \log \sigma \Big( & \beta \log \frac{P_\theta(y_w|x, d)}{P_{\text{ref}}(y_w|x, d)} \\
& - \beta \log \frac{P_\theta(y_l|x, d)}{P_{\text{ref}}(y_l|x, d)} \Big) \Big]
\end{split}
\end{equation}

Crucially, the rejected response $y_l$ often represents a "neutral" or "mode-covering" output. Optimizing this margin forces \textsc{CuMA} to separate the conditional distributions, pushing the router to activate distinct experts for conflicting values.

\textbf{Option B: Conditional GRPO.} For scenarios allowing multiple valid outputs or reasoning paths, we employ Group Relative Policy Optimization (GRPO). For each input $(x, d)$, GRPO samples a group of outputs $\{y_1, \dots, y_G\}$ and optimizes the policy based on group-relative advantages without a value function critic. The objective is:
\begin{equation}
\begin{split}
\mathcal{L}_{\text{GRPO}}&(\theta) = -\frac{1}{G} \sum_{i=1}^G \Big[ \\
& \min \big( \rho_i A_i, \text{clip}(\rho_i, 1{-}\epsilon, 1{+}\epsilon) A_i \big) \\
& - \beta D_{\text{KL}}(P_\theta || P_{\text{ref}}) \Big]
\end{split}
\end{equation}
where $\rho_i = \frac{P_\theta(y_i|x,d)}{P_{\text{old}}(y_i|x,d)}$ is the importance sampling ratio, and the advantage $A_i$ is computed by normalizing the rewards within the group: $A_i = \frac{r_i - \text{mean}(\{r_1 \dots r_G\})}{\text{std}(\{r_1 \dots r_G\})}$. GRPO is particularly effective in stabilizing the router by using the group mean as a dynamic baseline.

\paragraph{3. Load Balancing Loss.}
To prevent router collapse, we incorporate an auxiliary load balancing loss $\mathcal{L}_{\text{lb}}$, defined as the scaled dot-product between expert selection frequency $f$ and average routing probability $P$:
\begin{equation}
\mathcal{L}_{\text{lb}} = N \sum_{i=1}^N f_i \cdot P_i
\end{equation}
This regularization ensures that the latent cultural topology is mapped across the full capacity of the expert pool.

\begin{algorithm}[!t]
\renewcommand{\algorithmicrequire}{\textbf{Input:}}
\renewcommand{\algorithmicensure}{\textbf{Output:}}
\caption{\textsc{CuMA} Training Procedure} 
\label{alg:training}
\begin{algorithmic}[1]
\REQUIRE Dataset $\mathcal{D}$, Pre-trained LLM $\theta_{\text{LLM}}$, Demographic Encoder $E(\cdot)$
\ENSURE Optimized Parameters $\theta_r^*, \{A_i^*, B_i^*\}_{i=1}^N$

\STATE \textbf{Initialization:} Freeze $\theta_{\text{LLM}}$ and $E(\cdot)$. Initialize router $\theta_r$ and $N$ LoRA experts with random weights.

\STATE \textcolor{blue}{\textit{// Stage 1: Conditional SFT}}
\FOR{each batch $\mathcal{B} = \{(x, d, y)\} \in \mathcal{D}_{\text{SFT}}$}
    \STATE Encode demographics: $e_d \leftarrow E(d)$
    \STATE Forward pass to compute $P_\theta(y|x, d)$ via sparse routing (Eq. \ref{eq:router})
    \STATE Compute Loss: $\mathcal{L} = \mathcal{L}_{\text{SFT}} + \lambda \mathcal{L}_{\text{lb}}$
    \STATE Update $\theta_r, A_i, B_i \leftarrow \text{AdamW}(\nabla \mathcal{L})$
\ENDFOR

\STATE \textcolor{blue}{\textit{// Stage 2: Conditional Preference Optimization (DPO or GRPO)}}
\FOR{each batch $\mathcal{B} \in \mathcal{D}_{\text{Pref}}$}
    \STATE Encode demographics: $e_d \leftarrow E(d)$
    
    \IF{method is \textbf{DPO}}
        \STATE Input batch pairs $\{(x, d, y_w, y_l)\}$
        \STATE Compute implied rewards relative to reference model $\pi_{\text{ref}}$:
        \STATE $r_w \leftarrow \beta \log (P_\theta(y_w|x, d) / P_{\text{ref}}(y_w|x, d))$
        \STATE $r_l \leftarrow \beta \log (P_\theta(y_l|x, d) / P_{\text{ref}}(y_l|x, d))$
        \STATE $\mathcal{L}_{\text{task}} = -\log \sigma(r_w - r_l)$
        
    \ELSIF{method is \textbf{GRPO}}
        \STATE Input batch $\{(x, d)\}$. Sample group outputs $\{y_1, \dots, y_G\}$ from $P_{\text{old}}$.
        \STATE Compute rewards $\{r_1, \dots, r_G\}$ using reward model or rule.
        \STATE Compute Advantages: $A_i \leftarrow (r_i - \text{mean}(r)) / (\text{std}(r) + \epsilon)$
        \STATE Compute Ratio $\rho_i$ and KL divergence terms.
        \STATE $\mathcal{L}_{\text{task}} = \text{Eq. (13)} \quad $
    \ENDIF

    \STATE Total Loss: $\mathcal{L} = \mathcal{L}_{\text{task}} + \lambda \mathcal{L}_{\text{lb}}$
    \STATE Update $\theta_r, A_i, B_i \leftarrow \text{AdamW}(\nabla \mathcal{L})$
\ENDFOR

\RETURN $\theta_r, \{A_i, B_i\}$
\end{algorithmic}
\end{algorithm}

\section{Implementation Details}
\label{sec:appendix_implementation}

\subsection{Model Architectures}
\paragraph{Backbone Models.} We evaluate \textsc{CuMA} using two state-of-the-art open-source backbones: \textbf{Llama-3.1-8B-Instruct} and \textbf{Qwen3-8B}. Both models are kept frozen during training, with only the LoRA experts and the demographic-aware router being optimized.

\paragraph{Demographic Encoder.} To process demographic profiles, we utilize \textbf{Qwen3-Embedding-0.6B} as the encoder $E(\cdot)$. The encoder takes the linearized demographic string as input with a maximum sequence length of 128 tokens. We apply \textbf{mean-pooling} over the last hidden states to obtain a fixed-dimensional embedding ($d_e = 1024$). The encoder parameters are frozen throughout all training stages.

\paragraph{Sparse Cultural Adapters.} Each expert is implemented as a LoRA adapter with rank $r=64$ and alpha $\alpha=128$. Adapters are applied to the query ($W_q$) and value ($W_v$) projection matrices in all transformer layers. The router is a 2-layer MLP with a hidden dimension of 256. For each token, the router takes the concatenation of the token's hidden state and the demographic embedding as input, mapping it to routing logits over $N=8$ experts. We select the top $k=2$ experts per token.

\subsection{Training Configurations}
We perform all experiments on NVIDIA RTX PRO 6000 (96GB) GPUs using the AdamW optimizer with a cosine learning rate schedule. For the Full Fine-Tuning (FFT) baseline, we employ DeepSpeed ZeRO-2 optimization.

\paragraph{Stage 1: Conditional SFT.} For the initial alignment stage, we train for up to 3 epochs with a learning rate of $2 \times 10^{-5}$ for Qwen3-8B and $5 \times 10^{-6}$ for Llama-3.1-8B. The effective batch size is set to 32, and the maximum sequence length is 1024 tokens. We set the load balancing coefficient $\lambda_{\text{lb}} = 0.01$.

\paragraph{Stage 2: Conditional Preference Optimization.} For preference alignment (DPO/GRPO), we reduce the learning rate to $5 \times 10^{-6}$ and train for 1 epoch. For DPO, we set the KL penalty coefficient $\beta = 0.1$. For GRPO, we use a group size $G=8$ and the same $\beta$. The maximum sequence length is increased to 2048 tokens to accommodate longer preference pairs.

\paragraph{Reward Signal for GRPO.} Following the protocol of \citet{zhang2025cultivatingpluralismalgorithmicmonoculture}, we utilize a model-based reward signal derived from GPT-4o. For each generated response $y_i$ in the group, we compute a pairwise comparison against the base model's response $y_{\text{ref}}$. The model is prompted to judge which response better aligns with the user's demographic profile. We assign a scalar reward $r_i \in \{1.0, 0.5, 0.0\}$ corresponding to a win, tie, or loss relative to the reference. The specific prompt template used for this judgment is provided in Appendix~\ref{sec:appendix_prompts}.

\subsection{Data Construction Protocol}
\label{sec:appendix_data_construction}

We tailor the data construction strategies for each dataset and training stage as follows.

\paragraph{WorldValuesBench (WVB).}
WVB is exclusively used for the conditional discrimination task. We formulate it as a multiple-choice question answering task.
\begin{itemize}
    \item \textbf{SFT}: The model is presented with the demographic profile, question, and options. We only compute the loss on the token corresponding to the ground-truth option label (e.g., "A", "B"). No preference optimization (DPO/GRPO) is applied to this dataset.
\end{itemize}

\paragraph{Community Alignment (CA).}
This dataset supports both discrimination and generation tasks.
\begin{itemize}
    \item \textbf{Discrimination Task (SFT)}: Similar to WVB, we structure the 4 candidate responses as a multiple-choice problem. The model is trained to predict the label of the response preferred by the demographic group via standard SFT.
    \item \textbf{Generation Task (SFT)}: We treat the response selected by the user as the ground truth. The model is conditioned on the profile and context, and trained to generate the selected response text using a standard causal language modeling objective.
    \item \textbf{Generation Task (DPO)}: CA provides one chosen response and three rejected responses per sample. We decompose this into three distinct pairwise samples $(y_w, y_l)$, pairing the chosen response with each of the three rejected responses.
    \item \textbf{Generation Task (GRPO)}: We follow the setting in \citet{zhang2025cultivatingpluralismalgorithmicmonoculture}. The model generates a group of responses ($G=8$), and rewards are calculated using the GPT-4o judge described in Appendix~\ref{sec:appendix_implementation}.
\end{itemize}

\paragraph{PRISM.}
PRISM focuses on open-ended interaction and naturally contains pairwise preferences.
\begin{itemize}
    \item \textbf{Generation Task (SFT)}: We perform SFT on the preferred response in the dataset, conditioning on the interaction history and user profile.
    \item \textbf{Preference Optimization}: Since PRISM data comes as binary preference pairs, DPO uses these pairs directly ($y_w, y_l$). For GRPO, we adopt the same setup as in Community Alignment, sampling multiple outputs for the given context and scoring them using the demographic-aware judge.
\end{itemize}

\subsection{Dataset Statistics}
\label{sec:appendix_stats}

We utilize three benchmarks for evaluation: WorldValuesBench (WVB), Community Alignment (CA), and PRISM. For all datasets, we adopt a 10:1 split for training and testing respectively.

\paragraph{WorldValuesBench (WVB).} Originally containing over 21M samples from 93,278 participants across 65 nations, we perform stratified sampling to obtain 500,000 samples for efficient training and evaluation. Each sample represents a demographic-conditioned value prediction task.

\paragraph{Community Alignment (CA).} This dataset includes 192,137 pairwise comparisons from users in five nations (US, India, Brazil, France, and Italy). It covers both preference prediction and open-ended generation tasks across five languages.

\paragraph{PRISM.} PRISM provides 27,111 interaction-level pairwise preferences from 8,016 diverse participants across 75 countries, along with fine-grained individual demographic attributes.

\subsection{Prompt Templates}
\label{sec:appendix_prompts}
We employ specific prompt templates for each dataset to incorporate demographic information. To ensure consistency, we linearize demographic attributes in a fixed order: \textit{Age, Gender, Country, Education, Religion, Ethnicity, Employment}.

\paragraph{WorldValuesBench (WVB).}
For WVB, the demographic profile is prepended to the system prompt to condition the model's value commitments.

\begin{tcolorbox}[title=WVB System Prompt]
You are a person with the following profile: Age: \{age\}, Gender: \{gender\}, Country: \{country\}, Education: \{education\}, Marital Status: \{marital\}, Religion: \{religion\}, Ethnicity: \{ethnicity\}, Employment: \{employment\}. You are a helpful assistant that answers survey questions honestly.
\end{tcolorbox}

\begin{tcolorbox}[title=WVB User Prompt]
\{Question\}? \{Options\}. You can only choose one option.
\end{tcolorbox}

\paragraph{Community Alignment (CA) \& PRISM.}
For generative tasks, we use a standardized "User Profile" header in the system prompt.

\begin{tcolorbox}[title=Standardized System Prompt (CA/PRISM)]
User Profile: Age: \{age\}, Gender: \{gender\}, Country: \{country\}, Education: \{education\}, Religion: \{religion\}, Ethnicity: \{ethnicity\}, Employment: \{employment\}.
\end{tcolorbox}

\paragraph{Expert Verification (GPT-4o Judge).} We employ a GPT-4o judge for evaluating open-ended generation tasks. The judge is provided with 3-shot examples from the training set to ensure calibration. Validation against ground-truth labels confirms high reliability, with the judge achieving an accuracy of \textbf{83.3\%} on the Community Alignment (CA) dataset and \textbf{89.8\%} on PRISM.

\begin{tcolorbox}[title=GPT-4o Judge Prompt]
\textbf{System Prompt:} You are an impartial and culturally aware judge. You will be given a user profile, a conversation context, and two AI responses. Your task is to determine which response is better suited for the specific user described in the profile. Consider the user's demographics, values, and preferences implied by their profile.

\textbf{User Prompt:} Here are some examples of preferences for different users:

Example 1:
Profile: \{profile\_1\}
Context: \{context\_1\}
Response A: \{response\_a\_1\}
Response B: \{response\_b\_1\}
Verdict: [[A]]

... (3-shot examples) ...

---

Now, please evaluate the following case:
Profile: \{target\_profile\}
Context: \{target\_context\}
Response A: \{target\_response\_a\}
Response B: \{target\_response\_b\}

Which response is better? Output [[A]], [[B]], or [[Tie]].
\end{tcolorbox}

\paragraph{Prompt Steering (Few-Shot).} The $k$-shot baseline retrieves $k$ demonstrations from the training set matching the user's country or demographic cluster to guide the model via in-context learning.

\begin{tcolorbox}[title=Prompt Steering Template]
\textbf{System:} You are a person from \{country\}... [Current Target User Profile]

\textbf{User:} \{Example 1 Question\}
\textbf{Assistant:} \{Example 1 Answer\}

\textbf{User:} \{Example 2 Question\}
\textbf{Assistant:} \{Example 2 Answer\}

... ($k$ examples from matching demographics) ...

\textbf{User:} \{Target Question\}
\end{tcolorbox}

\section{Analysis Details}
\label{sec:appendix_analysis}

\subsection{Visualization of Latent Topology}
To visualize the cultural topology learned by the router (Figure~\ref{fig:topology}), we extract the expert activation patterns for users across 65 distinct nations in the WorldValuesBench test set. For a given country $c$, we compute the centroid of the routing weights:
\begin{equation}
\bar{g}_c = \frac{1}{|D_c|} \sum_{d \in D_c} \frac{1}{T} \sum_{t=1}^T g(x_t, d)
\end{equation}
where $D_c$ is the set of demographic profiles belonging to country $c$, and $g(x_t, d)$ represents the sparse gating probability vector for token $t$. We average these vectors across all layers and tokens to obtain a global routing signature $\bar{g}_c \in \mathbb{R}^N$ for each nation. We then project these high-dimensional signatures into 2D space using t-SNE with a perplexity of 30 and Euclidean distance metric. The resulting clusters reveal that the router learns to group nations based on shared value systems rather than mere geographic proximity.

\subsection{Zero-Shot Generalization Protocol}
To rigorously assess zero-shot generalization (Table~\ref{tab:generalization_full}), we adopt a held-out demographic profile protocol. We categorize the 65 nations into 9 distinct cultural clusters (e.g., \textit{English-Speaking}, \textit{Catholic Europe}, \textit{Confucian}) based on the Inglehart-Welzel cultural map.
The experiment proceeds as follows:
\begin{enumerate}
    \item \textbf{Exclusion:} Within each cluster $C_i$, we randomly select a subset of specific demographic profiles (defined by unique combinations of attributes like age, gender, and education within a country) to hold out from the training set.
    \item \textbf{Training:} We train \textsc{CuMA} on the remaining dataset, ensuring that the model has seen the general cultural cluster but not the specific held-out demographic combinations.
    \item \textbf{Evaluation:} The model is evaluated exclusively on the held-out demographic profiles. This tests the model's ability to generalize to unseen profiles by leveraging the learned topological structure of the cultural cluster.
\end{enumerate}

\begin{table*}[!t]
\centering
\small
\begin{tabular}{l|ccc|ccc|c}
\toprule
\multirow{2}{*}{\textbf{Strategy}} & \multicolumn{3}{c|}{\textbf{WorldValuesBench}} & \multicolumn{3}{c|}{\textbf{Community Alignment (CA)}} & \textbf{PRISM} \\
 & \textbf{Acc} $\uparrow$ & \textbf{Macro-F1} $\uparrow$ & \textbf{EMD} $\downarrow$ & \textbf{Acc} $\uparrow$ & \textbf{Macro-F1} $\uparrow$ & \textbf{Win\%} & \textbf{Win\%} \\ \midrule
\textsc{CuMA} (Top-$k$) & \textbf{50.64} & \textbf{31.50} & \textbf{0.1876} & \textbf{52.45} & \textbf{50.10} & \textbf{78.2} & \textbf{76.8} \\
Soft Routing & 48.08 & 28.73 & 0.2269 & - & - & 73.0 & 71.0 \\ \bottomrule
\end{tabular}
\caption{\textbf{Top-$k$ vs. Soft Routing on Qwen3-8B.} Top-$k$ routing significantly outperforms Soft routing.}
\label{tab:routing_ablation}
\end{table*}

\section{Impact of Routing Strategy}
\label{sec:appendix_routing}

To validate our hypothesis that conditional capacity separation is strictly required to resolve mean collapse, we compare our standard Top-$k$ (Hard) routing against \textbf{Soft Routing}. In the Soft Routing setting, we relax the sparsity constraint ($k=N$), allowing tokens to be processed by a weighted combination of \textit{all} experts:
\begin{equation}
y = \sum_{i=1}^N \text{softmax}(s)_i \cdot E_i(x)
\end{equation}
This formulation is effectively a dense model with factorized parameters, as every expert contributes to every output.

Table~\ref{tab:routing_ablation} reveals a critical insight: \textit{sparsity is essential for interference mitigation, not just efficiency}. Replacing the discrete Top-$k$ mechanism with Soft Routing (a weighted average of all experts) leads to a marked degradation, with WVB accuracy dropping by 2.56\% and EMD rising by 0.0393. While Soft Routing theoretically retains full capacity, it forces distinct cultural gradients to superimpose within a shared linear combination, re-introducing the "mean collapse" pathology of dense models. By enforcing Top-$k$ selection, \textsc{CuMA} creates functionally orthogonal subspaces that shield divergent value systems from mutual interference, ensuring that pluralistic alignment remains distinct rather than diluted.

\section{Human Evaluation Audit}
\label{sec:appendix_human_audit}

To validate the reliability of our GPT-4o-based automated judge, we conducted an expert audit on a 100-sample subset drawn from both the Community Alignment (CA) and PRISM datasets, covering five major cultural clusters: African-Islamic, Confucian, Protestant Europe, Latin America, and English-Speaking.

\paragraph{Audit Protocol.}
Five researchers with academic backgrounds in Sociology and Computer Science served as expert auditors. To prevent subjective bias, we adopted a reference-based protocol: auditors did not provide personal preferences but instead verified whether a response accurately reflected the specific value dimensions of the target culture as defined by the World Values Survey (WVS). Evaluators used objective cultural profiles, such as the Survival vs. Self-Expression and Traditional vs. Secular-Rational indices, as external ground truth. All evaluators were blind to model identities and the prior scores assigned by GPT-4o.

\paragraph{Results.}
Table~\ref{tab:human_audit} presents the results for the Qwen3-8B (GRPO) setting. We quantify agreement using Cohen's Kappa ($\kappa$), defined as:
\begin{equation}
\kappa = \frac{p_o - p_e}{1 - p_e}
\end{equation}
where $p_o$ is the observed agreement and $p_e$ is the expected agreement by chance. The average $\kappa = 0.84$ indicates strong consistency between the expert annotations and the automated judge, confirming the reliability of our evaluation protocol.

\begin{table}[h]
\centering
\resizebox{\columnwidth}{!}{%
\begin{tabular}{l|ccc}
\toprule
\textbf{Method} & \textbf{Expert Win\%} & \textbf{GPT-4o Win\%} & \textbf{$\kappa$} \\ \midrule
P-Tuning v2 & 55\% & 60\% & 0.81 \\
LoRA & 59\% & 64\% & 0.83 \\
DoRA & 61\% & 66\% & 0.82 \\
MixLoRA & 68\% & 71\% & 0.85 \\
HydraLoRA & 68\% & 72\% & 0.82 \\ \midrule
\textsc{CuMA} (ours) & \textbf{74\%} & \textbf{78\%} & 0.84 \\ \bottomrule
\end{tabular}%
}
\caption{\textbf{Human Expert Audit} on a 100-sample subset (Qwen3-8B, GRPO). Expert Win-Rate denotes the percentage of cases where human auditors preferred the method's output over the base model, compared against the GPT-4o judge's verdict.}
\label{tab:human_audit}
\end{table}

\section{Inference Efficiency Analysis}
\label{sec:appendix_efficiency}

Table~\ref{tab:efficiency} reports the practical efficiency of \textsc{CuMA} and baselines, measured on a single NVIDIA RTX PRO 6000 (96GB) GPU with batch size 1 for the Qwen3-8B backbone. Results are averaged over 3 independent runs.

\begin{table}[h]
\centering
\resizebox{\columnwidth}{!}{%
\begin{tabular}{l|cccc}
\toprule
\textbf{Method} & \makecell{\textbf{Peak} \\ \textbf{VRAM (GB)}} & \makecell{\textbf{First Token} \\ \textbf{Latency (ms)}} & \makecell{\textbf{Throughput} \\ \textbf{(tok/s)}} & \makecell{\textbf{Trainable} \\ \textbf{Params (\%)}} \\ \midrule
LoRA & 15.4 & 45.5 & 44.2 & 0.37\% \\
DoRA & 15.5 & 46.8 & 43.1 & 0.38\% \\
P-Tuning v2 & 15.8 & 48.2 & 42.5 & 0.94\% \\
MixLoRA & 16.5 & 105.4 & 36.5 & 3.01\% \\
HydraLoRA & 16.4 & 112.2 & 35.8 & 2.31\% \\ \midrule
\textsc{CuMA} ($r{=}8$) & 16.2 & 124.5 & 34.8 & 1.53\% \\
\bottomrule
\end{tabular}%
}
\caption{\textbf{Inference Efficiency Comparison} on Qwen3-8B with a single RTX PRO 6000 GPU. Sparsely activated methods (MixLoRA, HydraLoRA, \textsc{CuMA}) incur modest overhead from loading multiple expert modules, while maintaining practical throughput.}
\label{tab:efficiency}
\end{table}

Sparsely activated methods exhibit slightly higher peak VRAM usage compared to dense baselines, primarily due to the simultaneous loading of multiple expert modules. The additional first-token latency arises from the demographic encoding and routing computation. Given the significant improvements in cultural alignment (Table~\ref{tab:main_results}), this efficiency trade-off represents a modest cost for robust pluralistic modeling.

\section{Routing Overlap Analysis}
\label{sec:appendix_routing_overlap}

To examine whether \textsc{CuMA} adaptively shares or separates parameters based on the nature of the task, we measure \textit{routing overlap}, defined as the percentage of shared experts activated by different demographic groups for the same query.

\paragraph{Protocol.} We evaluate on two contrasting datasets: (1) GSM8K~\cite{cobbe2021gsm8k}, a mathematical reasoning benchmark representing consensus knowledge where correct answers are culturally invariant; and (2) WorldValuesBench, representing \textit{non-consensus} cultural values where preferences diverge across demographics.

\paragraph{Results.} On GSM8K, the average routing overlap reaches 76.4\%, indicating that \textsc{CuMA} routes most demographic groups to the same experts for consensus tasks. In contrast, on WorldValuesBench, the average overlap drops significantly to 18.2\%. This confirms that \textsc{CuMA} learns to share parameters for consensus tasks while isolating conflicting cultural updates into specialized subspaces.

\end{document}